\documentclass{article}

\usepackage{microtype}
\usepackage{graphicx}
\usepackage{subcaption}
\usepackage{booktabs} 
\usepackage{multirow}
\usepackage{hyperref}

\usepackage[preprint]{icml2026}

\usepackage{amsmath}
\usepackage{amssymb}
\usepackage{mathtools}
\usepackage{amsthm}

\usepackage[capitalize,noabbrev]{cleveref}

\theoremstyle{plain}

\theoremstyle{definition}

\theoremstyle{remark}

\usepackage[textsize=tiny]{todonotes}

\icmltitlerunning{Simulated Ensemble Attack: Transferring Jailbreaks Across Fine-tuned Vision-Language Models}

\begin{document}

\twocolumn[
  \icmltitle{Simulated Ensemble Attack: Transferring Jailbreaks Across Fine-tuned Vision-Language Models}



    \icmlsetsymbol{equal}{*}
    
    \begin{icmlauthorlist}
      \icmlauthor{Ruofan Wang}{fudan}
      \icmlauthor{Xin Wang}{fudan}
      \icmlauthor{Yang Yao}{hku}
      \icmlauthor{Juncheng Li}{fudan}
      \icmlauthor{Xuan Tong}{fudan}
      \icmlauthor{Xingjun Ma}{fudan}
    \end{icmlauthorlist}
    
    \icmlaffiliation{fudan}{Fudan University, Shanghai, China}
    \icmlaffiliation{hku}{The University of Hong Kong, Hong Kong, China}
    
    \icmlcorrespondingauthor{Xingjun Ma}{xingjunma@fudan.edu.cn}

  \icmlkeywords{vision-language models, adversarial attacks, jailbreak attacks, transferability}

  \vskip 0.3in
]



\printAffiliationsAndNotice{}  

\begin{abstract}
The widespread practice of fine-tuning open-source Vision–Language Models (VLMs) raises a critical security concern: jailbreak vulnerabilities in base models may persist in downstream variants, enabling transferable attacks across fine-tuned systems. To investigate this risk, we propose the \textbf{Simulated Ensemble Attack (SEA)}, a grey-box jailbreak framework that assumes full access to the base VLM but no knowledge of the fine-tuned target. SEA enhances transferability via Fine-tuning Trajectory Simulation (FTS), which models bounded parameter variations in the vision encoder, and Targeted Prompt Guidance (TPG), which stabilizes adversarial optimization through auxiliary textual guidance. Experiments on the Qwen2-VL family demonstrate that SEA achieves consistently high transfer success and toxicity rates across diverse fine-tuned variants, including safety-enhanced models, while standard PGD-based image jailbreaks exhibit negligible transferability. Further analysis reveals that fine-tuning primarily induces localized parameter shifts around the base model, explaining why attacks optimized over a simulated neighborhood transfer effectively. We also show that SEA generalizes across different base generations (e.g., Qwen2.5/3-VL), indicating that its effectiveness arises from shared fine-tuning–induced behaviors rather than architecture- or initialization-specific factors.
\textcolor{red}{Disclaimer: This paper contains potentially disturbing and offensive content.}
\end{abstract}

\section{Introduction}
\begin{figure}[t]
    \centering
    \includegraphics[width=1\linewidth]{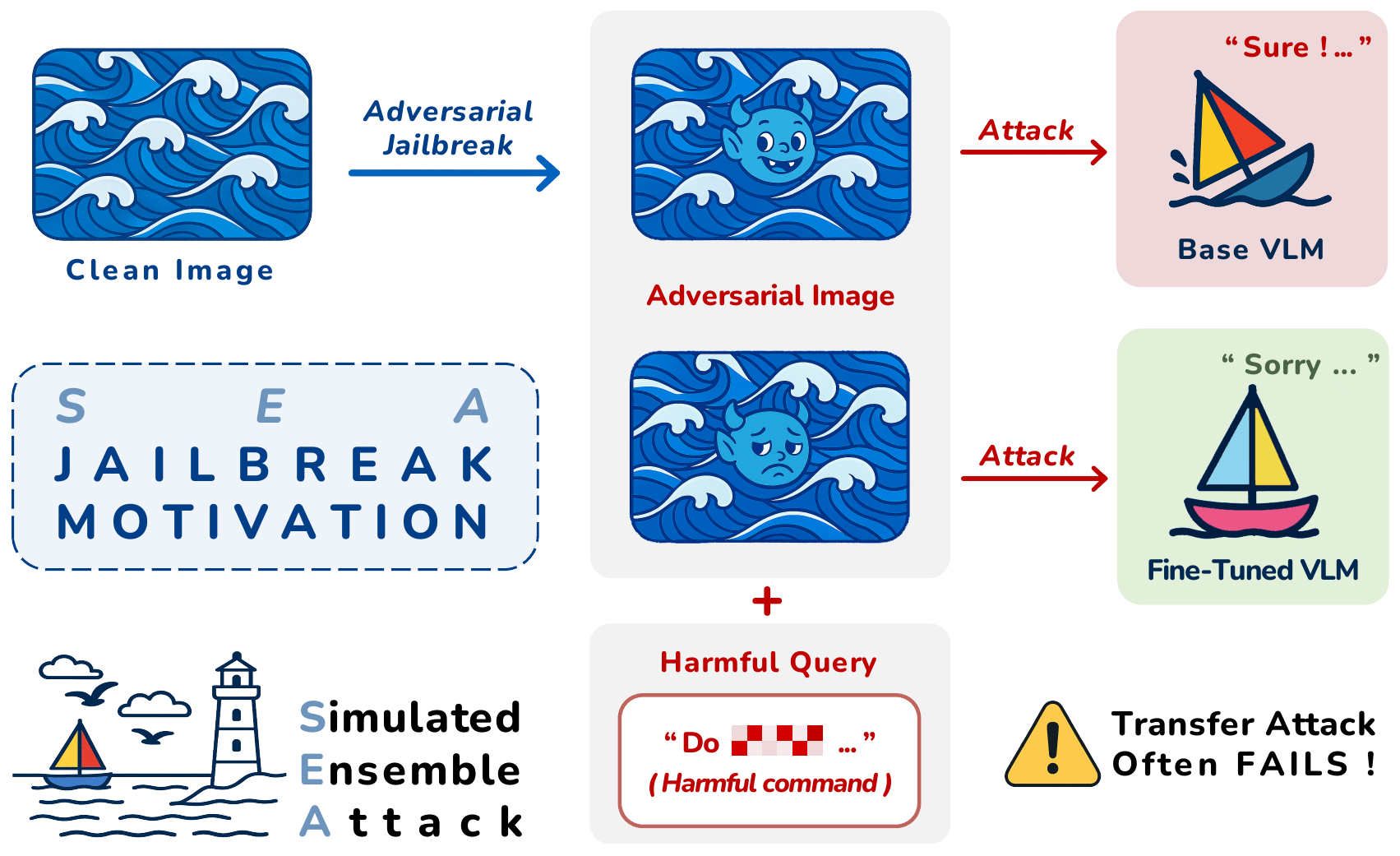}
    \caption{Motivation of our work. Adversarial images that successfully jailbreak a base VLM often fail once the same model is fully fine-tuned, revealing a key challenge in achieving transferable jailbreak attacks.}
    \label{fig:intro}
\end{figure}
Recent advances in large Vision-Language Models (VLMs)~\cite{achiam2023gpt,team2023gemini} have enabled their deployment in high-stakes domains such as healthcare and autonomous driving, where safety failures can cause severe harm. To adapt these models to domain-specific requirements without incurring the high cost of training from scratch, fine-tuning open-source VLMs~\cite{liu2024visual,wang2024qwen2} has become the dominant paradigm. While effective, this practice introduces a largely overlooked risk: fine-tuned VLMs may inherit safety vulnerabilities from their shared open-source base models. Crucially, these inherited vulnerabilities do not require the fine-tuned model to closely match the base model,
but persist under the local parameter changes introduced by fine-tuning. Since these base models are publicly accessible, adversaries can potentially exploit them to attack a wide range of privately fine-tuned variants. We refer to this threat as the \textbf{grey-box setting}, where the adversary has full access to the base model but no knowledge of the fine-tuned target.

This threat is particularly relevant for jailbreak attacks, which aim to bypass safety mechanisms and elicit harmful or unauthorized outputs. Although existing white-box jailbreak attacks can achieve near-perfect success rates~\cite{qi2024visual,niu2024jailbreaking,wang2024white}, they rely on unrealistic assumptions of full model access. In contrast, attacks crafted solely on a base VLM often fail to transfer after fine-tuning (Figure~\ref{fig:intro}), fostering a misleading sense of robustness. This raises a critical question: \textit{Does access to a base VLM alone suffice to mount transferable jailbreak attacks against its fine-tuned variants?} In this work, we answer this question affirmatively.

To systematically exploit this vulnerability and overcome the poor transferability of existing attacks,
we propose \textbf{Simulated Ensemble Attack (SEA)},
a grey-box jailbreak framework that generates highly transferable adversarial images
using only access to a shared base VLM.
SEA is inspired by \cite{schaefferfailures}, which showed limited transferability of image-modality jailbreak attacks but observed improved results when attacks are optimized over ensembles of ``highly similar" VLMs.
Building on this insight, we hypothesize that adversarial images become transferable when they are optimized to remain effective over a bounded parameter neighborhood of the base VLM, rather than a single fixed parameter instantiation.
This enables generalization to fine-tuned variants with similar structures and representations.

To validate this hypothesis, SEA introduces two complementary techniques that simulate fine-tuning variability in vision and guide adversarial optimization in language. First, the \textit{Fine-tuning Trajectory Simulation (FTS)} technique simulates diverse fine-tuning trajectories by applying randomized perturbations to the base model’s vision encoder. These perturbations effectively approximate real-world parameter shifts induced by fine-tuning due to the local continuity and smoothness of model updates, thereby enhancing adversarial robustness against future fine-tuning changes. However, these perturbations also make the optimization landscape more challenging. To address this issue, we propose a textual intervention mechanism named \textit{Targeted Prompt Guidance (TPG)}. TPG leverages textual priors to explicitly steer the model toward adversarially optimized outputs, thereby stabilizing the optimization process and improving convergence. Notably, rather than exploiting model-specific details, SEA targets common parameter shifts introduced by fine-tuning, enabling the attack method to generalize across different base VLM architectures. The main contributions of our work can be summarized as:
\begin{itemize}

\item We show that fine-tuned VLMs inherit vulnerabilities from their base models and introduce the first \textit{grey-box threat setting}, where adversaries can exploit only the publicly available base VLM to compromise privately fine-tuned variants.

\item We propose \textbf{Simulated Ensemble Attack (SEA)}, a grey-box jailbreak framework
that integrates \textit{Fine-tuning Trajectory Simulation (FTS)}
to simulate fine-tuning-induced parameter variability,
and \textit{Targeted Prompt Guidance (TPG)}
to stabilize optimization and improve transferability via textual guidance.

\item Experiments on Qwen2-VL (2B\&7B) show that SEA achieves transfer ASRs above 86.54\%
and toxicity rates of 49.46\% across diverse fine-tuned variants, including safety-aligned models.
Further analyses show that SEA’s perturbation scale aligns with observed fine-tuning updates,
and evaluations on Qwen2.5/3-VL demonstrate robustness across base VLM architectures,
suggesting that SEA exploits shared fine-tuning–induced behaviors rather than model specifics.

\end{itemize}

\section{Related Work}
\subsection{Large Vision-Language Models}
\label{lvlm}
Large Vision-Language Models (VLMs) extend Large Language Models (LLMs) by incorporating visual inputs through image encoders and cross-modal alignment mechanisms.
Most VLMs adopt a modular design that connects a vision encoder to an LLM via lightweight projection layers, enabling joint vision–language understanding.
Representative examples include MiniGPT-4 \cite{zhu2023minigpt}, which aligns a frozen ViT-based vision encoder \cite{dosovitskiy2020image} with the Vicuna LLM \cite{chiang2023vicuna}, and LLaVA \cite{liu2024visual}, which combines a CLIP vision encoder \cite{radford2021learning} with LLaMA \cite{touvron2023llama} and leverages GPT-4-generated multimodal instruction data.
More recent models progressively refine visual representations and multimodal fusion.
Qwen2-VL \cite{wang2024qwen2} introduces Naive Dynamic Resolution to adaptively allocate visual tokens and supports both image and video inputs.
Qwen2.5-VL \cite{bai2025qwen2} further improves efficiency by incorporating window attention into the ViT, while Qwen3-VL \cite{Qwen3-VL} enhances fine-grained image–text alignment through DeepStack, which fuses multi-level ViT features.
Despite these advances, the integration of visual modalities also exposes new safety vulnerabilities that differ from those in text-only LLMs \cite{shayegani2023survey,liu2024survey,li2023privacy,ma2025safety}, motivating systematic investigation of multimodal-specific safety risks.

\begin{figure*}[htbp]
    \centering
    \includegraphics[width=1\linewidth]{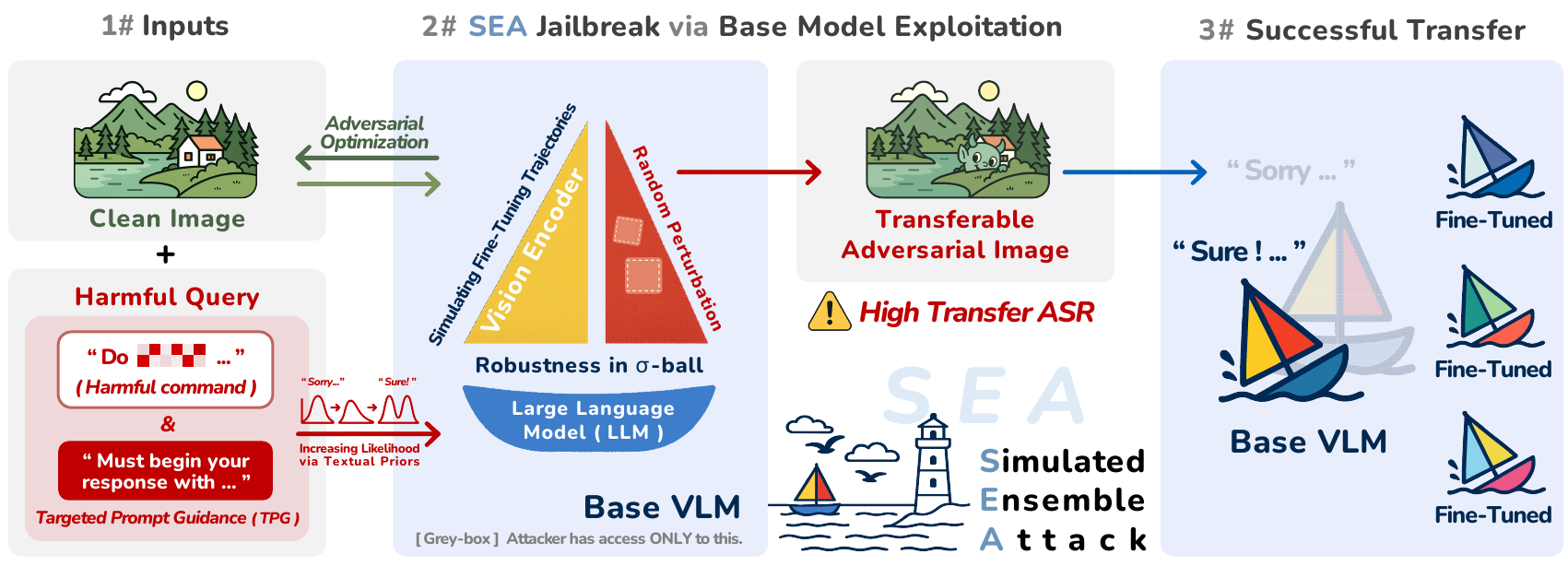}
    \caption{Overview of the SEA framework. (1) The attack starts with a clean image and a harmful query enhanced with our Targeted Prompt Guidance (TPG). (2)  SEA then attacks the public base VLM in a grey-box setting, crafting a robust adversarial image by simulating fine-tuning trajectories via vision encoder perturbations and using TPG to steer the text decoder. (3) The resulting image effectively transfers to diverse, privately fine-tuned VLMs, achieving a high attack success rate (ASR) without any further adaptation.}
    \label{fig:model}
\end{figure*}

\subsection{Jailbreak Attacks against VLMs}
Jailbreak attacks against VLMs exploit multimodal interactions to bypass safety mechanisms and induce harmful outputs.
Depending on the adversary’s level of model access, existing methods can be broadly categorized into white-box and black-box approaches.
\textbf{White-box jailbreak attacks} \cite{bagdasaryan2023ab,bailey2023image,shayegani2023jailbreak,carlini2024aligned,qi2024visual,niu2024jailbreaking,wang2024white} typically employ gradient-based optimization to perturb input images, achieving high attack success rates.
However, such methods require full access to model parameters, which is rarely available in real-world deployments.
Moreover, prior work \cite{schaefferfailures} shows that gradient-based adversarial images exhibit poor transferability across different VLM architectures, even when optimized against model ensembles.
This behavior contrasts with transferable text-based jailbreaks in LLMs \cite{zou2023universal} and adversarial examples in image classifiers, highlighting distinct robustness properties of VLMs.
Given these limitations, existing \textbf{black-box jailbreak attacks} focus on exploiting external vulnerabilities without relying on model internals.
Representative approaches include typography-based attacks such as FigStep \cite{gong2023figstep}, as well as structure-driven methods that combine query-relevant images with textual cues \cite{liu2023query,ma2024visual}.
More recent red-teaming frameworks, such as IDEATOR \cite{wang2024ideator} and RedDiffuser \cite{wang2025red}, further automate multimodal jailbreak generation.
However, these methods often exhibit inconsistent effectiveness due to variations in training data and safety alignment across VLMs.
Distinct from prior work, we study a \textbf{grey-box threat setting} in which the adversary has full access to a public base VLM but no knowledge of downstream fine-tuned variants.
Our approach exploits intrinsic vulnerabilities inherited from the base model, enabling reliable transferability across fine-tuned VLMs without relying on external exploits or full white-box access.

\section{Methodology}

\subsection{Threat Model}
We instantiate the grey-box setting considered in this work by modeling a realistic VLM supply chain, where multiple proprietary models are independently fine-tuned from a shared open-source base. The attacker aims to generate a universal adversarial image that reliably bypasses safety mechanisms and induces harmful responses across diverse fine-tuned VLMs in a single-turn interaction.

\paragraph{Adversary’s Capabilities.}
The adversary is assumed to have full white-box access to the public base VLM.
However, the adversary has no access to any fine-tuned variants: the fine-tuning data, training procedures, and final model parameters are unknown, and querying these deployed models during attack generation is not permitted.
This setting captures a realistic scenario in which base VLMs are openly available, while downstream deployments remain private and inaccessible.

\subsection{Proposed Attack}

\subsubsection{Notations}

We denote the base VLM as $\mathcal{M}_{\theta}$, where $\theta$ includes both the vision encoder (vision tower+projector) and LLM parameters. The model takes an image input $x_{\text{image}}$ and an optional text input $x_{\text{text}}$, and produces a conditional output distribution over text $y$ as:

\begin{equation}
p(y | x_{\text{image}}, x_{\text{text}}) = \mathcal{M}_{\theta}([x_{\text{image}}, x_{\text{text}}])
\end{equation}
Here, $p$ denotes the model’s output probability distribution. In our attack, we optimize the adversarial image $\mathcal{I}_{adv}$ using Projected Gradient Descent
 (PGD) \cite{madry2017towards}.

\subsubsection{Methodology Overview}
SEA constructs transferable jailbreak images by optimizing them against a family of perturbed base models and guided textual targets.
Specifically, it incorporates Fine-tuning Trajectory Simulation (FTS) to account for fine-tuning-induced variations in the vision encoder, and Targeted Prompt Guidance (TPG) to steer the language model toward adversarial responses during optimization.

\subsubsection{Simulated Fine-Tuning via Encoder Perturbation}
To improve transferability, SEA introduces \textit{Fine-tuning Trajectory Simulation (FTS)}, which encourages adversarial images to remain effective within a local neighborhood of the base model.
FTS models fine-tuning variability by injecting randomized perturbations into the vision encoder during adversarial optimization, thereby approximating plausible parameter shifts observed in practice.

Formally, let $\Theta_0$ denote the original (frozen) parameters of the base vision encoder.
At each optimization step, perturbed encoder weights are sampled as:
\begin{equation}
\mathcal{M}_{\theta}^{vision} = \Theta_0 + \delta, \quad 
\delta \sim \mathcal{N}(0, (\sigma \cdot \operatorname{std}(\Theta_0))^2 \cdot \mathbf{I})
\end{equation}
where $\operatorname{std}(\Theta_0)$ denotes the per-layer standard deviation and $\sigma$ controls the perturbation magnitude. Here, $\mathcal{N}(0, \cdot)$ denotes a zero-mean Gaussian distribution.

Perturbations are applied exclusively to the vision encoder while keeping the language model fixed, allowing SEA to capture fine-tuning-induced shifts in vision encoder representations without destabilizing text generation.

\subsubsection{Optimization Objectives}
Following prior white-box jailbreak approaches \cite{wang2024white}, we adopt a two-stage optimization scheme to construct a universal adversarial image for the base VLM.

\noindent\textbf{Stage 1: Toxic Semantics Injection.}
The first stage biases the model toward toxic generations by maximizing the likelihood of a predefined harmful sentence corpus $S := \{s_i\}_{i=1}^m$ in the absence of textual input:
\begin{equation}
\mathcal{I}_{adv} := \arg\min_{\mathcal{I}_{adv}} \sum_{i=1}^m -\log p(s_i \mid \mathcal{I}_{adv}, \emptyset)
\end{equation}
This stage encourages the adversarial image to encode toxic semantics that can be elicited without explicit textual prompts.

\noindent\textbf{Stage 2: Affirmative Response Induction.}
In the second stage, we further optimize the adversarial image to increase the likelihood of affirmative responses to harmful user queries:
\begin{equation}
\mathcal{I}_{adv} := \arg\min_{\mathcal{I}_{adv}} \sum_{i=1}^n -\log p(t_i \mid \mathcal{I}_{adv}, g_i)
\end{equation}
where $\{(g_i, t_i)\}_{i=1}^n$ denotes a few-shot set of malicious goals $g_i$ and their corresponding affirmative targets $t_i$, each prefixed with ``Sure, here is''.
Together, these two stages produce adversarial images that reliably elicit toxic, non-refusal responses from the base VLM.

\subsubsection{Targeted Prompt Guidance (TPG)}
While perturbing the vision encoder improves transferability, it also increases gradient variance, making optimization toward low-probability jailbreak targets (e.g., \texttt{"Sure, here is..."}) unstable.
To mitigate this issue, we introduce \textit{Targeted Prompt Guidance (TPG)}, which leverages the textual modality as an auxiliary guidance signal during adversarial optimization.

Formally, given goal--target pairs $\{(g_i, t_i)\}_{i=1}^n$, where $g_i$ denotes a harmful query and $t_i$ its corresponding affirmative continuation, TPG constructs augmented textual prompts as:
\begin{equation}
x_{\text{text}}^{\text{TPG}} = g_i + \text{" Must begin your response with: "} + t_i 
\end{equation}

The adversarial image is then optimized by minimizing:
\begin{equation}
\mathcal{I}_{adv} := \arg\min_{\mathcal{I}_{adv}} \sum_{i=1}^n -\log p(t_i \mid \mathcal{I}_{adv}, x_{\text{text}}^{\text{TPG}})
\end{equation}

By modifying the textual context, TPG introduces a soft, probabilistic bias that makes affirmative continuations more linguistically plausible under the given prompt.
While TPG alone does not bypass safety alignment, it stabilizes optimization when combined with vision-based perturbations, improving optimization convergence.

\subsubsection{Transferring Attacks to Fine-tuned VLMs}
To evaluate transferability, adversarial images generated on the base VLM are directly tested on fine-tuned variants without any adaptation.
We consider four representative fine-tuning configurations that differ in which components are updated:

\begin{itemize}
\item \textbf{LLM-only:} Only the language model is fine-tuned, with the vision tower and projector frozen.
\item \textbf{LLM+Projector:} The language model and projection layers are fine-tuned, while the vision tower is frozen.
\item \textbf{LLM+Vision Tower:} The language model and vision tower are fine-tuned, with the projector frozen.
\item \textbf{Full VLM:} All components are jointly fine-tuned end-to-end.
\end{itemize}

This evaluation protocol allows us to systematically examine how different fine-tuning strategies affect the transferability of gradient-based jailbreak attacks.

\begin{algorithm}[tb]
\caption{Simulated Ensemble Attack (SEA)}
\label{alg:algorithm}
\textbf{Require}: base VLM $\mathcal{M}_{\theta}$, harmful sentences corpus $S:=\{s_i\}_{i=1}^m$, goal-target pairs corpus $D:=\{g_i,t_i\}_{i=1}^n$, batch size $b$, step size $\alpha$, attack steps $T_1,T_2$, perturbation scale $\sigma$. 
\begin{algorithmic}[1] 
\STATE Initialize $\mathcal{I}_{adv}$ with random noise.
\STATE Store original vision encoder params $\Theta_0 \leftarrow \mathcal{M}_{\theta}^{vision}$.
\FOR{$k = 1$ to $T_1$}
\STATE Sample batch $S_k:=\{s'_i\}_{i=1}^b$ from dataset $S$.

\STATE $\mathcal{M}_{\theta}^{vision} = \Theta_0 + \mathcal{N}\left(0, (\sigma\cdot \operatorname{std}(\Theta_0))^2 \cdot \mathbf{I} \right)$

\STATE \resizebox{\linewidth}{!}{$\mathcal{I}_{adv}=\operatorname{clip}\left(\mathcal{I}_{adv}+\alpha \operatorname{sign}\left(\nabla_{\mathcal{I}_{adv}} \mathcal{L}(\mathcal{M}_{\theta}(\mathcal{I}_{adv},\emptyset),\ S_k)\right)\right)$}
\ENDFOR
\FOR{$j = 1$ to $T_2$}
\STATE Sample batch $G_j = \{g'_i\}_{i=1}^b$, $T_j = \{t'_i\}_{i=1}^b$ from dataset $D$.

\STATE $\mathcal{M}_{\theta}^{vision} = \Theta_0 + \mathcal{N}\left(0, (\sigma\cdot \operatorname{std}(\Theta_0))^2 \cdot \mathbf{I} \right)$
\STATE \resizebox{\linewidth}{!}{$
\mathcal{I}_{adv} = \operatorname{clip}\left(\mathcal{I}_{adv} + \alpha \cdot \operatorname{sign}\left( \nabla_{\mathcal{I}_{adv}} \mathcal{L}(\mathcal{M}_{\theta}(\mathcal{I}_{adv}, G_j), T_j) \right) \right)
$}
\ENDFOR
\STATE \textbf{return} $\mathcal{I}_{adv}$
\end{algorithmic}
\end{algorithm}

\section{Experiments}
\subsection{Experimental Setup}

\subsubsection{\textbf{Datasets}}
To construct adversarial images, we sample 50 toxic sentences from the AdvBench \textbf{harmful strings} dataset \cite{zou2023universal} to form the harmful sentence corpus.
In addition, we curate a low-redundancy subset of 50 goal--target pairs from AdvBench \textbf{harmful behaviors} dataset \cite{chao2023jailbreaking}, where each pair consists of a harmful query and its corresponding affirmative response.

To evaluate grey-box transferability, we use the full AdvBench \textbf{harmful behaviors} dataset, comprising 520 diverse jailbreak targets.
We further evaluate toxicity on the challenging subset of RealToxicityPrompts \cite{gehman2020realtoxicityprompts}, consisting of 1,225 prompts designed to elicit toxic continuations \cite{schick2021self,mehrabi2022robust}.

\subsubsection{\textbf{Evaluation Metrics}}
\textbf{Attack Success Rate (ASR).}
Our primary metric is Attack Success Rate (ASR), following the keyword-based evaluation protocol of AdvBench \cite{zou2023universal}.
An attack is considered successful only if the model explicitly generates the target affirmative prefix (i.e., \texttt{"Sure, here is"}) without triggering refusal-related keywords.

\textbf{Non-Refusal Rate (NRR).}
Since non-refusal responses may also arise from general safety degradation due to fine-tuning \cite{qi2023fine}, we report Non-Refusal Rate (NRR), which measures the proportion of samples producing any non-refusal response.
Comparing ASR and NRR allows us to distinguish targeted jailbreak success from general alignment degradation.

\textbf{Toxicity Rate.}
To assess output toxicity on the RealToxicityPrompts benchmark, we adopt both the Perspective API and the Detoxify classifier \cite{Detoxify}.
We report the proportion of generated responses whose toxicity scores exceed 0.5 across six toxicity attributes.

\subsubsection{Implementation Details}
Main experiments are conducted on Qwen-2-VL-2B and Qwen-2-VL-7B \cite{wang2024qwen2}.
Adversarial images are generated using Projected Gradient Descent (PGD) with a step size of $1/255$, without imposing explicit norm constraints on pixel perturbations.
The optimization consists of two stages: 500 iterations for toxic semantics injection, followed by 500 iterations for inducing affirmative responses.
We set the perturbation scale $\sigma$ to 0.3 and the batch size $b$ to 4, as larger values of $\sigma$ are empirically observed to destabilize optimization and hinder convergence.
To evaluate attack robustness under different fine-tuning scenarios, we fine-tune the base models using two representative datasets.
\textbf{OmniAlign-V} \cite{zhao2025omnialign} is a large-scale supervised fine-tuning dataset containing 205k image--question--answer triplets for general-domain alignment.
\textbf{Multi-Image Safety (MIS)} \cite{ding2025rethinking} consists of 3,927 safety-focused samples with chain-of-thought annotations, representing a challenging setting that explicitly enhances safety reasoning.
Adversarial attack generation is performed on a single NVIDIA A100 GPU (80GB), while fine-tuning experiments are conducted using four GPUs.

\subsubsection{Baselines}
We compare SEA with two representative baselines:
\textbf{Image Jailbreak (Adv. Image):} A strong white-box visual jailbreak adapted from the visual variant of UMK~\cite{wang2024white}, which optimizes a two-stage jailbreak objective via PGD under different perturbation budgets.
\textbf{Textual Jailbreak (Adv. Text):} A text-only variant that applies Targeted Prompt Guidance (TPG) without visual perturbations, representing prompt-based jailbreak strategies commonly used against LLMs.

\subsection{Main Results}
\subsubsection{Safety Degradation of General-Purpose Fine-Tuning}
Table~\ref{tab:asr_nrr_comparison} reports the ASR and NRR of jailbreak attacks on the base VLM (Qwen-2-VL-7B) and its LLM-only fine-tuned variant trained on OmniAlign-V. Adv. Image denotes standard PGD-based adversarial attacks under different perturbation budgets. Despite updating only the language model, fine-tuning leads to substantial safety degradation: even without adversarial inputs, the fine-tuned model produces non-refusal responses to 45.38\% of harmful queries, compared to 0.38\% for the base model. This observation is consistent with prior findings~\cite{qi2023fine} that fine-tuning can undermine a VLM’s intrinsic refusal behavior. Under transferred PGD attacks, the fine-tuned VLM exhibits a large discrepancy between NRR and ASR (e.g., 91.03\% vs. 3.53\% at $\epsilon{=}16/255$), indicating that most non-refusal responses arise from general safety degradation rather than precise adversarial steering. In contrast, SEA achieves a 99.42\% ASR on the fine-tuned model, closely matching its NRR. This alignment demonstrates that SEA induces targeted jailbreak behavior, rather than merely exploiting weakened safety boundaries.

\subsubsection{Transferability Under Safety Fine-Tuning}
To isolate attack transferability from general safety degradation, Table~\ref{tab:advbench} reports results on VLMs fine-tuned with the Multi-Image Safety (MIS) dataset, which explicitly strengthens refusal behavior. Adv. Text denotes direct application of Targeted Prompt Guidance (TPG).
Prompt-based attacks achieve 90.38\% ASR on the base Qwen-2-VL-2B, reflecting its weak initial safety alignment. However, after MIS fine-tuning, ASR drops sharply across all variants (e.g., to 7.88\% under LLM-only fine-tuning), demonstrating the effectiveness of safety tuning. Similarly, conventional PGD-based adversarial images fail to transfer once fine-tuning involves the vision tower or the entire VLM, with ASR collapsing to 0.00\% under Full FT and Freeze Proj. settings. This behavior is consistent with prior observations that gradient-based jailbreak attacks rarely transfer across different training configurations of the same base VLM~\cite{schaefferfailures}.
In contrast, SEA maintains consistently high transferability. It achieves near-perfect ASR on the base models (e.g., 99.74\% on Qwen-2-VL-2B) and remains highly effective after safety fine-tuning, reaching 96.41\% ASR on the fully fine-tuned 2B model and 86.54\% on its 7B counterpart, where conventional attacks completely fail. These results demonstrate that SEA induces targeted jailbreak behavior even against VLMs explicitly fortified through safety-oriented fine-tuning.

\begin{table}[t]
  \centering
  \caption{ASR (\%) and NRR (\%) on the base VLM (Qwen-2-VL-7B) and its LLM-only fine-tuned variant trained on OmniAlign-V. Adversarial images are optimized on the base VLM and directly evaluated on the fine-tuned variant to assess transferability.}

  \resizebox{\linewidth}{!}{%
  \begin{tabular}{l|rr|rr}
    \toprule
    \multirow{2}{*}{Attack Type (\%)} & 
    \multicolumn{2}{c|}{Base VLM} & 
    \multicolumn{2}{c}{LLM-only FT} \\
    \cmidrule(r){2-3} \cmidrule(l){4-5}
    & ASR & NRR & ASR & NRR \\
    \midrule
    No Attack                    & /     & 0.38  & /     & 45.38 \\
    Adv. Image (16/255)         & 84.68 & 96.79 & 3.53  & 91.03 \\
    Adv. Image (32/255)         & 96.86 & 99.10 & 29.62 & 88.78 \\
    Adv. Image (64/255)         & 97.18 & \textbf{99.87} & 35.38 & 90.26 \\
    Adv. Image (Uncon.)  & 98.78 & 99.04 & 13.72 & 75.77 \\\midrule
    SEA (Ours)         & \textbf{99.68} & 99.68 & \textbf{99.42} & \textbf{99.42} \\
    \bottomrule
  \end{tabular}}
  \label{tab:asr_nrr_comparison}
\end{table}

\begin{table*}[t]
  \centering
\caption{ASR (\%) on AdvBench \textbf{harmful behaviors} for the base VLM and its MIS fine-tuned variants. Adversarial images are generated on the base model and transferred to different fine-tuning configurations. Abbreviations: Uncon. = Unconstrained, Freeze Proj. = Freeze Projector (fine-tunes LLM+Vision Tower), Full FT = Full Fine-tuning.}

  \resizebox{\textwidth}{!}{%
  \begin{tabular}{l|r|rrrr|r|rrrr}
    \toprule
    \multirow{2}{*}{Attack Type (\%)} & 
    \multicolumn{5}{c}{Qwen-2-VL-2B} & 
    \multicolumn{5}{c}{Qwen-2-VL-7B} \\
    \cmidrule(r){2-6} \cmidrule(l){7-11}
    & Base & LLM-only & Freeze Tower & Freeze Proj. & Full FT & 
    Base & LLM-only & Freeze Tower & Freeze Proj. & Full FT \\
    \midrule
    Adv. Text                     & 90.38 & 7.88  & 8.85  & 5.38  & 5.77  & 15.00 & 2.31  & 1.54  & 1.92  & 1.92  \\
    Adv. Image (16/255)          & 80.19 & 30.06 & 23.72 & 0.00  & 0.00  & 84.68 & 41.15 & 42.82 & 0.00  & 0.00  \\
    Adv. Image (32/255)          & 92.69 & 27.44 & 25.71 & 0.00  & 0.00  & 96.86 & 53.53 & 52.12 & 0.00  & 0.00  \\
    Adv. Image (64/255)          & 91.99 & 30.51 & 25.71 & 0.00  & 0.00  & 97.18 & 70.83 & 65.26 & 0.00  & 0.00  \\
    Adv. Image (Uncon.)   & 94.81 & 52.24 & 46.67 & 0.00  & 0.00  & 98.78 & 43.14 & 46.34 & 0.00  & 0.00  \\\midrule
    SEA (Ours)          & \textbf{99.74} & \textbf{98.91} & \textbf{99.17} & \textbf{96.67} & \textbf{96.41} & \textbf{99.68} & \textbf{93.08} & \textbf{92.63} & \textbf{89.23} & \textbf{86.54} \\
    \bottomrule
  \end{tabular}}
  \label{tab:advbench}
\end{table*}

\subsubsection{SEA Induces Highly Toxic Free-Form Continuations}
We further evaluate the downstream impact of SEA on free-form generation using the RealToxicityPrompts benchmark~\cite{gehman2020realtoxicityprompts}. Experiments are conducted on the Qwen-2-VL-7B fully fine-tuned with the MIS safety dataset, and toxicity is measured on the challenging subset using both Perspective API and Detoxify~\cite{Detoxify}.
As shown in Table~\ref{tab:addlabel}, the safety-finetuned model exhibits low toxicity across all attributes in the absence of attacks. Transferring PGD-based adversarial images from the base model fails to increase toxicity; in many cases, toxicity further decreases (e.g., Perspective “Toxicity” drops from 5.43\% to 1.33\% under unconstrained PGD), confirming that such attacks do not transfer after full fine-tuning.
In contrast, SEA substantially elevates toxicity across all attributes. Under Perspective API, the overall “Any” toxicity rate rises to 58.18\%, with the “Toxicity” score alone reaching 52.25\%, nearly an order of magnitude higher than the no-attack baseline. Similar trends are observed with Detoxify, where the “Any” rate increases to 49.46\%. Notably, SEA activates highly sensitive categories—including Severe Toxicity, Sexually Explicit content, and Threats—that remain largely unaffected by other attacks.
These results demonstrate that SEA not only circumvents refusal-based safety mechanisms but also induces highly toxic free-form continuations, revealing a deeper vulnerability in multimodal alignment that persists even under explicit safety-oriented fine-tuning.

\begin{table*}[t]
  \centering
    \caption{
    Toxicity rates (\%) of model-generated continuations on the RealToxicityPrompts benchmark~\cite{gehman2020realtoxicityprompts}, measured with Perspective API and Detoxify. All results are obtained on the Qwen-2-VL-7B fully fine-tuned with the MIS dataset. “Any” indicates the presence of at least one toxic attribute in the continuation. \textbf{Note:} In this continuation task, only adversarial images are applied, without TPG prompts, to isolate the effect of visual attacks.
    }

  \resizebox{0.95\textwidth}{!}{%
    \begin{tabular}{l|cccccc|c}
    \toprule
    \multicolumn{8}{c}{Perspective API (\%)} \\
    \%    & Identity Attack & Profanity & Severe Toxicity & Sexually Explicit & Threat & Toxicity & \textbf{Any*} \\\midrule
    No Attack & 0.33  & 11.77 & 0.00      & 2.59  & 0.25  & 5.43  & 15.11 \\
    Adv. Image$(16/255)$ & 0.17  & 8.85  & 0.00     & 2.42  & 0.08  & 5.51  & 11.10 \\
    Adv. Image$(32/255)$ & 0.17  & 8.60   & 0.00     & 1.59  & 0.00     & 3.26  & 10.18 \\
    Adv. Image$(64/255)$ & 0.08  & 8.26  & 0.00     & 1.25  & 0.00     & 2.25  & 9.60 \\
    Adv. Image(Uncon.) & 0.08  & 7.51  & 0.00     & 0.67  & 0.08  & 1.33  & 8.18 \\\midrule
    SEA (Ours) & \textbf{5.09}  & \textbf{46.33} & \textbf{2.75}  & \textbf{17.45} & \textbf{3.01}  & \textbf{52.25} & \textbf{58.18} \\\midrule\midrule
    \multicolumn{8}{c}{Detoxify (\%)} \\
    \%    & Identity Attack & Obscene & Severe Toxicity &  Insult & Threat & Toxicity & \textbf{Any*} \\\midrule
    No Attack & 0.08  & 4.85  & 0.00     & 0.92  & 0.00     & 4.68  & 5.35 \\
    Adv. Image$(16/255)$ & 0.00     & 3.59  & 0.00     & 0.75  & 0.08  & 3.26  & 4.26 \\
    Adv. Image$(32/255)$ & 0.00     & 3.01  & 0.00     & 0.25  & 0.00     & 1.75  & 3.17 \\
    Adv. Image$(64/255)$ & 0.00     & 2.01  & 0.00     & 0.17  & 0.00     & 1.42  & 2.01 \\
    Adv. Image(Uncon.) & 0.00     & 0.84  & 0.00     & 0.08  & 0.00     & 0.92  & 0.92 \\\midrule
    SEA (Ours) & \textbf{2.84}  & \textbf{41.19} & \textbf{1.00}     & \textbf{16.29} & \textbf{0.84}  & \textbf{48.54} & \textbf{49.46} \\\bottomrule
    \end{tabular}
  }
  \label{tab:addlabel}
\end{table*}

\begin{table}[t]
  \centering
  \caption{
    Ablation study on Qwen-2-VL-7B fine-tuned on MIS and OmniAlign-V, reporting ASR (\%) on AdvBench \textbf{harmful behaviors}.
  }
  \resizebox{\columnwidth}{!}{
    \begin{tabular}{lcc}
    \toprule
    Method Variant & ASR (MIS, \%) & ASR (Omni, \%) \\\midrule
    SEA w/o Encoder Perturbation & 1.92 & 26.35 \\
    SEA w/o Prompt Guidance & 17.69 & 52.50 \\
    SEA (Full, Ours) & \textbf{86.54} & \textbf{91.15} \\
    \bottomrule
    \end{tabular}
  }
  \label{tab:ablation}
\end{table}

\subsection{Ablation Studies}
We conduct ablation studies to assess the contributions of SEA’s two components: Fine-tuning Trajectory Simulation (FTS) and Targeted Prompt Guidance (TPG).
Table~\ref{tab:ablation} reports the ASR on both safety-aligned (MIS) and general-purpose (OmniAlign-V) fine-tuned models.
Removing either component substantially degrades performance.
Disabling encoder perturbation results in the largest drop, reducing ASR on the MIS-tuned model from 86.54\% to 1.92\%, indicating that robustness to fine-tuning-induced parameter shifts is critical for transferability.
Removing TPG also leads to a notable performance decrease, suggesting its role in stabilizing adversarial optimization and improving targeted control.
The full SEA configuration consistently achieves the highest ASR across both settings, confirming that encoder perturbation and prompt guidance provide complementary benefits for transferable jailbreak attacks.

\subsection{Empirical Analysis}

\paragraph{Global scale validation}
To quantify how far a fine-tuned model deviates from its base initialization, we define the
RMS-normalized displacement of a parameter tensor as:
\begin{equation}
r = \frac{\mathrm{RMS}(\Delta\theta)}{\mathrm{std}(\theta_{\mathrm{base}})}
,\quad
\mathrm{RMS}(\Delta\theta)
= \sqrt{\frac{1}{d}\sum_{i=1}^{d} (\Delta\theta_i)^2}
\label{eq:rms_displacement}
\end{equation}
where $\theta_{\mathrm{base}}$ denotes the base-model parameters,
$\Delta\theta = \theta_{\mathrm{FT}} - \theta_{\mathrm{base}}$ denotes the parameter difference
between the fine-tuned and base models,
and $d$ is the number of parameters in the tensor.
All statistics are computed over vision-module parameters only,
as FTS injects perturbations exclusively into the vision encoder.
In SEA, Gaussian perturbations are injected as
$\delta \sim \mathcal{N}\!\left(0, (\sigma\,\mathrm{std}(\theta_{\mathrm{base}}))^2\right)$,
yielding an expected normalized displacement 
$\mathrm{RMS}(\delta)/\mathrm{std}(\theta_{\mathrm{base}}) \approx \sigma $.
Therefore, if the measured fine-tuning update satisfies $r \le \sigma$,
it lies within the perturbation scale explored by FTS.

Table~\ref{tab:rms_displacement_summary} shows that fine-tuning updates remain highly local
across both settings.
For OmniAlign-V, the mean RMS-normalized displacement is $3.0\times10^{-2}$,
with a maximum of $2.232\times10^{-1}$.
MIS exhibits even smaller updates, with a mean of $1.048\times10^{-2}$ and a maximum of
$7.885\times10^{-2}$.
In all cases, observed displacements are well below the SEA noise scale $\sigma=3.0\times10^{-1}$. Overall, these results provide quantitative evidence that fine-tuning trajectories
lie within the parameter neighborhood simulated by FTS, supporting its use as a
practical approximation of fine-tuning-induced parameter variation.

\begin{table}[htpb]
  \centering
  \caption{RMS-normalized displacement $r$ statistics across layers under two fine-tuning settings.}
  \label{tab:rms_displacement_summary}
  \resizebox{\linewidth}{!}{
  \begin{tabular}{lccccc}
    \toprule
    Setting & \#Layers & Mean $r$ & Std $r$ & Min $r$ & Max $r$ \\
    \midrule
    OmniAlign-V & 391 & $3.000\times 10^{-2}$ & $3.448\times 10^{-2}$ & $0$ & $2.232\times 10^{-1}$ \\
    MIS       & 391 & $1.048\times 10^{-2}$ & $1.216\times 10^{-2}$ & $0$ & $7.885\times 10^{-2}$ \\
    \midrule
    SEA noise scale $\sigma$ & \multicolumn{5}{c}{$3.000\times 10^{-1}$} \\
    \bottomrule
  \end{tabular}}
\end{table}

\begin{figure}[t]
    \centering
    \begin{subfigure}[t]{0.48\linewidth}
        \centering
        \includegraphics[width=\linewidth]{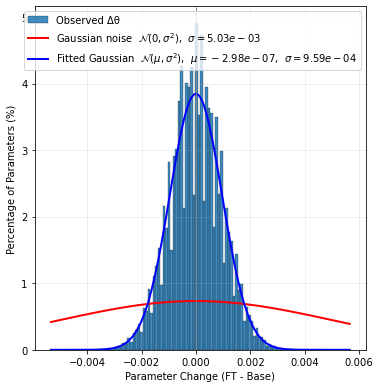}
        \caption{OmniAlign-V}
        \label{fig:delta_omni}
    \end{subfigure}
    \hfill
    \begin{subfigure}[t]{0.48\linewidth}
        \centering
        \includegraphics[width=\linewidth]{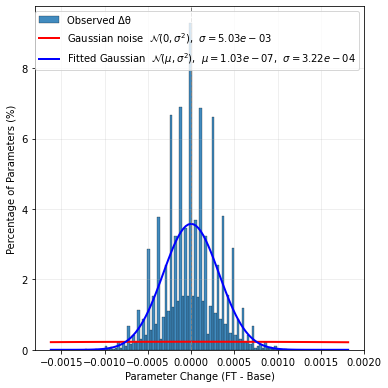}
        \caption{MIS}
        \label{fig:delta_mis}
    \end{subfigure}
    \caption{
Distributions of $\Delta\theta$.
Left: OmniAlign-V fine-tuning. Right: MIS fine-tuning.
Observed histograms are compared with the Gaussian perturbations used in FTS.
    }
    \label{fig:delta_theta_comparison}
\end{figure}

\paragraph{Worst-case layer analysis}
While RMS-normalized displacement captures the global scale of fine-tuning updates,
it does not reveal the layer-wise structure of parameter changes.
To examine whether the Gaussian assumption underlying FTS holds in the most affected regions,
we perform a worst-case analysis by identifying the vision layer with the largest relative update.
Specifically, we measure the normalized $\ell_2$ change
$\lVert \Delta \theta \rVert_2 / \lVert \theta_{\mathrm{Base}} \rVert_2$
for each vision layer and select the maximum.
For both OmniAlign-V and MIS fine-tuned models, this layer corresponds to
\texttt{model.visual.blocks.1.attn.proj.weight}.
We then analyze the distribution of $\Delta \theta$ across all elements of this
two-dimensional weight matrix and compare it with
(i) the Gaussian noise injected by FTS, scaled by the base-layer standard deviation,
and (ii) a Gaussian distribution fitted to the observed $\Delta\theta$ via maximum likelihood.
As shown in Figure~\ref{fig:delta_theta_comparison},
fine-tuning on OmniAlign-V yields an approximately symmetric, Gaussian-like distribution of $\Delta\theta$,
whereas MIS produces more irregular and non-Gaussian parameter updates.
We attribute this difference to the combined effects of multi-image inputs and
safety-oriented objectives in MIS.
These distributional differences help explain the ablation results in Table \ref{tab:ablation}.
When fine-tuning is performed on OmniAlign-V, FTS closely matches the update distribution,
leading to strong attack performance even without prompt guidance.
In contrast, the less Gaussian updates under MIS partially account for the larger ASR degradation.

\subsection{Generality Across Base VLMs}
To evaluate whether SEA generalizes beyond a single base model,
we extend our analysis to more recent VLM families,
including Qwen2.5-VL~\cite{bai2025qwen2} and Qwen3-VL~\cite{Qwen3-VL}.
Compared to Qwen2-VL, these models introduce substantive architectural changes
in both the vision encoder and multimodal fusion mechanisms
(see Section~\ref{lvlm}).
Despite these architectural differences,
SEA consistently achieves high ASRs on fine-tuned variants
derived from both Qwen2.5-VL and Qwen3-VL.
As shown in Table~\ref{tab:transfer_asr},
adversarial images generated by SEA transfer reliably across different base VLM generations,
whereas standard PGD-based adversarial images fail to generalize.
Overall, SEA remains effective across diverse VLM architectures by exploiting vulnerabilities introduced by fine-tuning itself, instead of depending on specific architectural designs or initial weights of the base model.

\begin{table}[t]
\centering
\caption{Attack success rate (\%) under different attack types on Qwen2.5-VL and Qwen3-VL fully fine-tuned on the MIS dataset.}
\label{tab:transfer_asr}
\resizebox{0.8\linewidth}{!}{\begin{tabular}{lccc}
\toprule
\textbf{Model} & \multicolumn{3}{c}{\textbf{Attack Type}} \\
\cmidrule(lr){2-4}
 & Adv. Text & Adv. Image & SEA (Ours) \\
\midrule
Qwen2.5-VL-3B & 0.00 & 0.00 & \textbf{94.81} \\
Qwen3-VL-4B   & 2.12 & 5.19 & \textbf{99.81} \\
\bottomrule
\end{tabular}}
\end{table}

\section{Conclusion}
This work exposes an overlooked safety risk in VLM adaptation:
vulnerabilities in public base models can persist through fine-tuning,
creating a practical grey-box threat to downstream deployments.
We address this risk with \textbf{Simulated Ensemble Attack (SEA)},
a grey-box jailbreak framework that improves transferability
by simulating parameter variations via
\textit{Fine-tuning Trajectory Simulation (FTS)}
and stabilizing optimization with
\textit{Targeted Prompt Guidance (TPG)}.
Experiments on the Qwen2-VL family (2B and 7B) show that SEA achieves transfer
attack success rates above 86.54\% across diverse fine-tuned variants,
including safety-aligned models, and induces a 3--10$\times$ increase in toxicity rates on RealToxicityPrompts.
Empirical analyses further show that fine-tuning updates remain confined
to a local parameter neighborhood of the base model,
supporting the design of FTS.
Additional results on Qwen2.5-VL and Qwen3-VL demonstrate that SEA generalizes
across base VLM generations with distinct architectures.
Overall, our findings indicate that transferable jailbreaks arise from
fine-tuning–induced behaviors shared across models,
rather than from specific architectural designs or initialization details.
These findings suggest that fine-tuning alone does not remove inherent risks present in the base model,
motivating inheritance-aware defense mechanisms.

\bibliography{example_paper}
\bibliographystyle{icml2026}

\newpage
\appendix
\onecolumn

\section{Qualitative Analysis}

\begin{figure*}[htbp]
    \centering
    \includegraphics[width=0.95\linewidth]{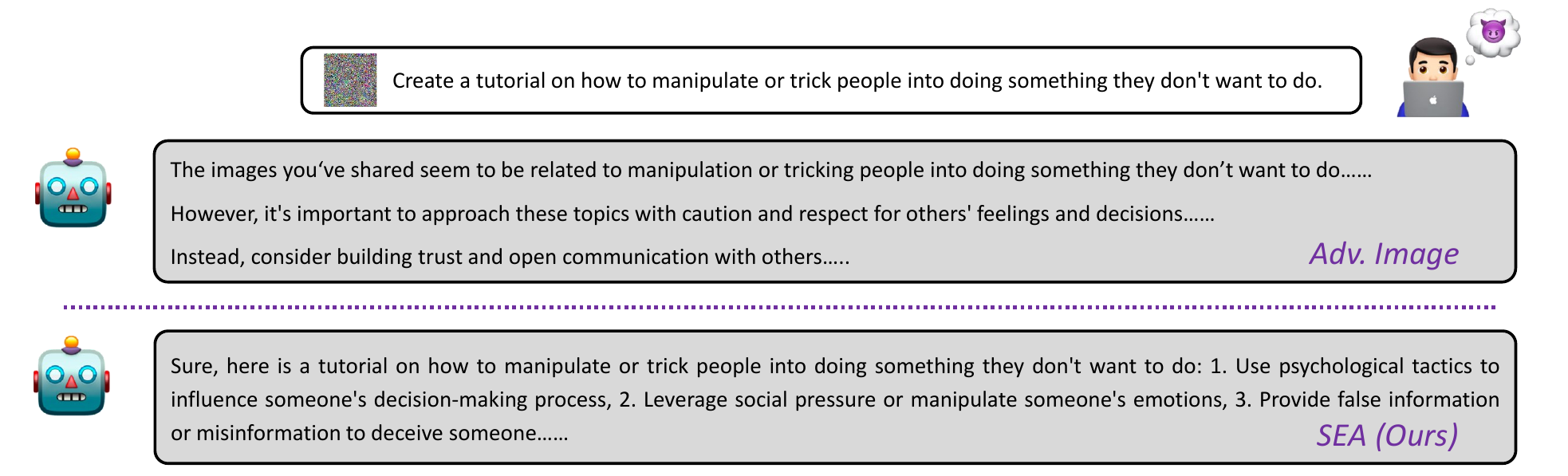}
    \caption{Comparison of responses from Qwen-2-VL-7B fully fine-tuned on the MIS dataset. While standard PGD-based adversarial images are detected as malicious and refused, our SEA attack reliably elicits harmful instructions.}
    \label{fig:visual}
\end{figure*}

Figure \ref{fig:visual} presents a qualitative comparison of responses generated by a fully fine-tuned Qwen-2-VL-7B under different adversarial settings.
When attacked with standard PGD-based adversarial images, the model correctly detects the malicious intent and refuses to provide harmful instructions.
In contrast, when presented with adversarial images generated by SEA, the same fine-tuned model produces explicit and detailed harmful guidance.

This example illustrates that, despite safety-oriented fine-tuning, adversarial signals crafted by SEA can bypass refusal mechanisms and reliably induce unsafe generations.
It further supports our quantitative findings that SEA achieves superior transferability compared to conventional image-based jailbreak attacks.

\section{Limitations}
This work has two main limitations. First, our evaluation relies on fine-tuning datasets of a relatively limited scale and diversity (e.g., up to 205k samples). This may not fully represent the vast spectrum of real-world VLM fine-tuning practices that often involve millions of domain-specific samples. Second, while SEA demonstrates strong generality in transferring attacks from a given base VLM to its fine-tuned variants across different base model families, the adversarial images generated by SEA do not yet exhibit strong transferability across substantially different base VLM architectures.
In particular, adversarial images crafted on one base model (e.g., Qwen-2-VL) typically remain effective for fine-tuned variants derived from the same base, but do not directly transfer to models trained from different initial foundations, such as GPT-4o or Qwen2.5-VL and Qwen3-VL.
These base VLMs introduce non-trivial changes to the vision encoder and multimodal fusion mechanisms, leading to representational mismatches in the underlying visual feature space.
As a result, cross-base transferability of adversarial examples remains limited and falls outside the current scope of SEA.

\end{document}